\documentclass[letterpaper]{article} 
\usepackage{aaai25}  
\usepackage{times}  
\usepackage{helvet}  
\usepackage{courier}  
\usepackage[hyphens]{url}  
\usepackage{graphicx} 
\usepackage{natbib}  
\usepackage{caption} 
\usepackage{algorithm}
\usepackage{algorithmic}

\usepackage{subcaption}
\usepackage{amsmath}
\usepackage{amsfonts}
\usepackage{xargs}
\usepackage{bm}
\usepackage{booktabs}
\usepackage{multirow}
\usepackage{amssymb}

\newcommand{\bfd}{\mathbf{d}}

\newcommand{\bfh}{\mathbf{h}}
\newcommand{\bfi}{\mathbf{i}}

\newcommand{\bfr}{\mathbf{r}}

\newcommand{\bfv}{\mathbf{v}}

\newcommand{\bfE}{\mathbf{E}}

\newcommand{\bfR}{\mathbf{R}}

\newcommand{\bfV}{\mathbf{V}}
\newcommand{\bfW}{\mathbf{W}}

\newcommand{\rmP}{\mathrm{P}}
\newcommand{\rmQ}{\mathrm{Q}}

\newcommand{\rmS}{\mathrm{S}}

\newcommand{\calC}{\mathcal{C}}

\newcommand{\calE}{\mathcal{E}}

\newcommand{\calG}{\mathcal{G}}
\newcommand{\calH}{\mathcal{H}}

\newcommand{\calK}{\mathcal{K}}
\newcommand{\calL}{\mathcal{L}}

\newcommand{\calN}{\mathcal{N}}

\newcommand{\calR}{\mathcal{R}}

\newcommand{\calU}{\mathcal{U}}
\newcommand{\calV}{\mathcal{V}}

\newcommand{\bbR}{\mathbb{R}}

\newcommand{\bmPsi}{\bm{\Psi}}

\newcommandx{\norm}[3][2={}, 3={}]{\vert\vert #1 \vert\vert_{#2}^{#3}}

\usepackage{xcolor}

\newcommandx{\sumlim}[3][1={i}, 2={1}]{\sum_{#1 = #2}^{#3}}
\newcommandx{\prodlim}[3][1={i}, 2={1}]{\prod_{#1 = #2}^{#3}}

\DeclareMathOperator*{\argmax}{arg\,max}

\newcommand{\RNN}{\mathrm{RNN}}

\newcommand{\MLP}{\mathrm{MLP}}
\newcommand{\ud}{\mathrm{d}}
\newcommand{\uq}{\mathrm{q}}
\newcommand{\uk}{\mathrm{k}}
\newcommand{\uv}{\mathrm{v}}

\newcommand{\barh}{\bar{h}}
\newcommand{\bark}{\bar{k}}
\usepackage{newfloat}
\usepackage{listings}
\DeclareCaptionStyle{ruled}{labelfont=normalfont,labelsep=colon,strut=off} 
\floatstyle{ruled}
\newfloat{listing}{tb}{lst}{}
\floatname{listing}{Listing}
\title{Deep Representation Learning for Forecasting Recursive and Multi-Relational Events in Temporal Networks}

\author {
    Tony Gracious,  
    Ambedkar Dukkipati
}
\affiliations {
    Department of Computer Science and Automation\\
    Indian Institute of Science, Bangalore - 560012, INDIA \\
    \{tonygracious, ambedkar\}@iisc.ac.in
}

\usepackage{bibentry}

\begin{document}

\maketitle

\begin{abstract}
Understanding relations arising out of interactions among entities can be very difficult, and predicting them is even more challenging. This problem has many applications in various fields, such as financial networks and e-commerce. These relations can involve much more complexities than just involving more than two entities. One such scenario is evolving recursive relations between multiple entities, and so far, this is still an open problem. This work addresses the problem of forecasting higher-order interaction events that can be multi-relational and recursive. We pose the problem in the framework of representation learning of temporal hypergraphs 
that can capture complex relationships involving multiple entities. The proposed model, 
\textit{Relational Recursive Hyperedge Temporal Point Process} (RRHyperTPP) uses an encoder that learns a dynamic node representation based on the historical interaction patterns and then a hyperedge link prediction-based decoder to model the occurrence of interaction events. These learned representations are then used for downstream tasks involving forecasting the type and time of interactions. The main challenge in learning from hyperedge events is that the number of possible hyperedges grows exponentially with the number of nodes in the network. This will make the computation of negative log-likelihood of the temporal point process expensive, as the calculation of survival function requires a summation over all possible hyperedges. In our work, we develop a noise contrastive estimation method to learn the parameters of our model, and we have experimentally shown that our models perform better than previous state-of-the-art methods for interaction forecasting. 
\end{abstract}



\section{Introduction}
With the advancement of the field of geometric deep learning and the availability of more and more datasets that capture complex interactions among entities, there is a new interest in learning representations from higher-order relations in networks~\citep{Bronstein2017:EtAL:GeometricDeepLearningGoingBeyondEuclideanData,Bronstein2021:EtAL:GeometricDeepLearningGridsGroupsGraphsGeodesicsAndGauges}. Until now the focus had been only on hypergraphs, where relations are represented as hyperedges that represent interactions of groups of entities/nodes~\citep{GhoshdastidarDukkipati:2017:ConsistencyOfSpectralHypergraphPartitioningUnderPlantedPartitionModel,GhoshdastidarDukkipati:2017:UniformHypergraphPartitioning,LeeEtAL:2023:ImMeWeReUsAndImUsTridirectionalConstrastiveLearningOnHypergraphs}. With the availability of richer datasets that show more complex interaction events, learning problems must consider scenarios beyond hypergraphs by incorporating a wide variety of information associated with these interactions. These ideas have been explored in recent works of topology-aware deep learning, where higher-order network structures such as simplicial complexes, cell complexes, and multi-scale topological information are used for improving graph neural networks \citep{HajijEtAL:2023:TopologicalDeepLearningGoingBeyondGraphData, HornEtAL:2022:TopologicalGraphNeuralNetworks}. Recent works have shown that $n$-ary facts or multi-relational hyperedges are the natural data representations for Knowledge graphs, and models based on those perform better than that use pairwise multi-relation models~\citep{WenEtAL:2016:OnTheRepresentationAndEmbeddingOfKnowledgeBasesBeyondBinaryRelations,FatemiEtAL:2020:KnowledgeHypergraphsPredictionBeyondBinaryRelations}. 

Most of these works ignore the temporal aspects of the network, where entities interact and evolve~\citep{ShubhamEtAL:2019:AGenerativeModelForDynamicNetworksWithApplications}. This has been addressed using dynamic network models based on temporal point processes (TPP) and link prediction. The recent works model interactions as instantaneous edges \citep{KumarEtAL:2019:PredictingDynamicEmbeddingTrajectoryInTemporalInteractionNetworks,DaiEtAL:2016:DeepCoevolutionaryNetworkEmbeddingUserAndItemFeaturesForRecommendation,TrivediEtAL:2019:DyRepLearningRepresentationsoverDynamicGraphs,CaoEtAL:2021:DeepStructuralPointProcessForLearningTemporalInteractionNetworks}/hyperedges \citep{GraciousEtAL:2023:DynamicRepresentationLearningWithTemporalPointProcessesForHigherOrderInteractionForecasting} forming between a pair of nodes or a group of nodes, and a generative model is parameterized to model the formation of edges given the history. 

\begin{figure}
    \centering
    \includegraphics[width=\linewidth]{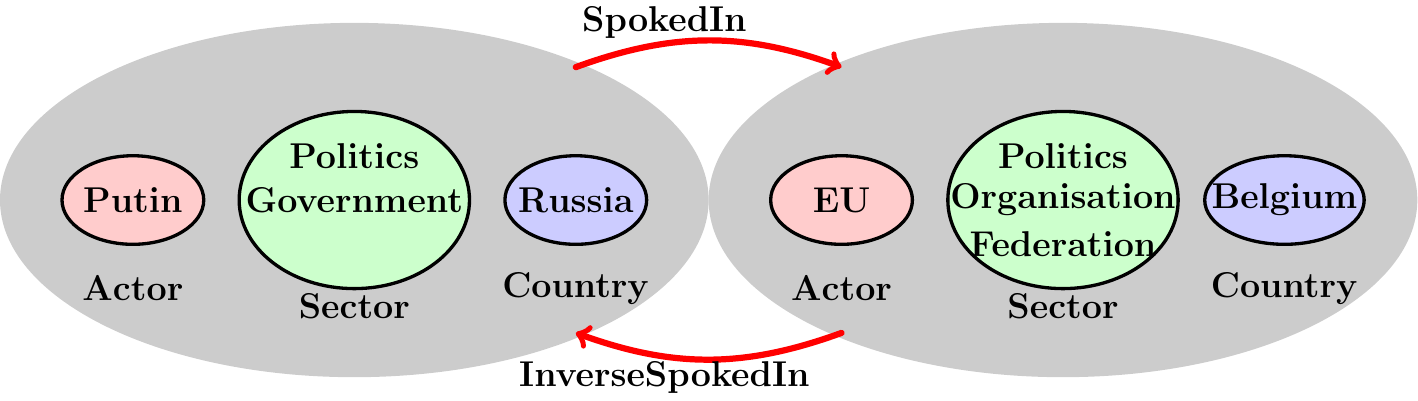}
     \caption{Real world events represented as Relational Recursive Hyperedges. Here, there are source and target hyperedges with three relations inside them and relation types between them to indicate the nature of the interaction.}
    \label{fig:Orderd_KG}
\end{figure}

However, real-world interactions can be much more complex than just a group of entities interacting. These interactions can exhibit group structure within themselves. For example, a directed hyperedge involves interaction between two groups of entities in the source and destination. Here, one can represent the source and destination as hyperedges, which act as nodes to the interaction hyperedge. In our work, we generalize this recursive group structure property of real-world interactions to incorporate a variable number of groupings. Examples of these types of interactions can be seen in political events, where source and target hyperedges function as nodes within larger hyperedges. Figure \ref{fig:Orderd_KG} illustrates a real-world event where \texttt{Putin} delivers a speech at the \texttt{EU}. In this context, each source and target hyperedge consists of nodes that are themselves hyperedges, representing the actor, sector, and country. The source and target hyperedges are connected by the relation \texttt{SpokedIn}, indicating the action of the source hyperedge when giving a speech, while the relation \texttt{InvSpokedIn} represents the inverse relation, with the target acting as the source and the source as the target. 

In this paper, we use a multi-relational recursive hyperedges structure to incorporate these complex higher-order interactions. Here, hyperedges can act as nodes in other hyperedges (like CCed or receiver groups in email interactions) and contain relation types indicating their role in the interactions. Then, a TPP is defined with these hyperedges as event types with conditional intensity function parameterized using representations/embeddings of the nodes and relations involved in the interactions.  Furthermore, the previous approaches in TPP based interaction forecasting use negative sampling to approximate the loss associated with survival function, which is intractable to calculate due to the huge number of possible event types in the TPP. This can introduce biases in training, especially in these complex interactions.  To address these challenges, we propose the model \textit{Relational Recursive Hyperedge Temporal Point Process} (RRHyperTPP). The following are the contributions of this work. 1)We propose a contrastive learning strategy for higher-order interaction for hyperedge events in networks. This provides a technique for training the model without maximum likelihood estimation, thereby avoiding the computationally expensive survival function calculation. We establish that the proposed model performs better than the previous models that use negative sampling; 2) We proposed a new deep learning architecture for learning node representation from higher-order interaction. This involves two stages: interaction update and temporal drift. Interaction update is used to revise the node representation when an event involving it occurs by using the features of the interaction. Temporal drift is used to model the  evolution of node representation during the interevent period using time projection based on JODIE~\citep{KumarEtAL:2019:PredictingDynamicEmbeddingTrajectoryInTemporalInteractionNetworks} or Neural ODE~\citep{ChenEtAL:2018:NeuralOdinaryDifferntialEquations} or time embeddings based on Fourier time features~\citep{XuEtAL:2020:InductiveRepresentationLearningOnTemporalGraphs};
 3) Finally, we propose a new method for hyperedge link prediction for multi-relational recursive hypergraphs; 4) Extensive experiments are done to show the advantage of the proposed method over existing state-of-the-art models.

\section{Related Works}
\label{sec:related_workds}

The recent works on modeling continuous time dynamic networks are of two kinds: those that are trained using TPP loss and those that use link prediction loss. The models in the first category parameterize the probability density function of a Neural Temporal Point Process with edges as the mark associated with events~\citep{Shchur:EtAL:2021:NeuralTemporalPointProcessesAReview}. These involve models DeepCoevolve \citep{DaiEtAL:2016:DeepCoevolutionaryNetworkEmbeddingUserAndItemFeaturesForRecommendation}, DyReP \citep{TrivediEtAL:2019:DyRepLearningRepresentationsoverDynamicGraphs}, DSPP \citep{CaoEtAL:2021:DeepStructuralPointProcessForLearningTemporalInteractionNetworks}, and GNPP \citep{XiaEtAL:2022:GraphNeuralPointProcessForTemporalInteractionPrediction} that uses neural networks to parameterize the conditional intensity function of a TPP based on the nodes involved in the edges. These works mostly focus on interaction forecasting on bipartite and homogeneous networks. More recently, GHNN~\citep{HanEtAL:2020:GraphHawkesNeuralNetworkForForecastingOnTemporalKnowledgeGraphs} defined TPPs on Temporal Knowledge Graphs where entities of different kinds and interactions are associated with relation type. However, all these models are developed for pair-wise interaction forecasting and cannot accurately model the higher-order interactions. Recent work \citep{GraciousEtAL:2023:DynamicRepresentationLearningWithTemporalPointProcessesForHigherOrderInteractionForecasting} modeled higher-order interactions as hyperedge event formation in a network. Here, they used recurrent neural network architecture to capture the evolution of nodes with each interaction and used fourier-based time features to model the dynamic of node representation during intervention time. These node representations are then used by a hyperedge-based link predictor to define the conditional intensity function of the TPP. Although this model can capture interactions between any number of nodes, it does not consider the internal groupings of nodes or their relationships with other nodes, such as in email exchanges or co-authorship networks.

The works in the second category use graph representation learning to learn continuous-time representations for the nodes that are then used to predict links using a binary scoring function. This involves models like JODIE \citep{KumarEtAL:2019:PredictingDynamicEmbeddingTrajectoryInTemporalInteractionNetworks}, TGAT \citep{XuEtAL:2020:InductiveRepresentationLearningOnTemporalGraphs}, and TGN \citep{RossiEtAL:2020:TemporalGraphNetworksForDeepLearningOnDynamicGraphs} which uses combination of temporal graph attention and recurrent neural networks to learn dynamic node representation. Recently, CAt-Walk \citep{BehrouzEtAL:2023:CATWalkInductiveHypergraphLearningViaSetWalks} have shown to predict higher-order interaction in the form of hyperedges. Here, they used causal
random walks, which maintain the temporal order of interactions to create features that are then used by a set of prediction-based architectures to forecast hyperedges as binary classification problems at the time of interest. However, all these models can only predict the presence of an interaction at a given time and cannot answer temporal queries like when a group of entities will interact or the time of the next interaction.

\section{Preliminaries and Problem Statement}
Tables \ref{tab: Notations} and \ref{tab: Notations_2} in Appendix \ref{sec:appendix:notations} summarize important notations used in this work.
 
\paragraph{Hypergraph.} A Hypergraph is denoted by the tuple $\calG = (\calV, \calH)$, where $\calV = \{v_1, v_2, \ldots, v_{|\calV|}\}$ is the set of nodes, $\calH$ is the set of  hyperedges with $\calH \subset \calC( \calV)$ and $\calC(\calV)$ is the powerset of $\calV$.
\paragraph{Recursive Hypergraph.} A Recursive Hypergraph $\calG = (\calV, \calH)$ is a generalization of Hypergraph, where hyperedges can contain nodes and other hyperedges.  Let, $ \calH^\ell = \calC( \cup_{i=0}^{\ell-1} \calH^i  ) $, if $\ell \geq 1$, where $\calH^0 = \calC( \calV )  $. A recursive hypergraph is of depth $\ell$ if $\calH \subset \calH^\ell$.  
This work focuses on recursive hypergraphs of depth one and two. 
%
\paragraph{Temporal Multi-relational Recursive Hypergraphs.}  A Temporal  Multi-relational Recursive hypergraph is denoted by tuple $\calG (t)= ( \calV, \calH, \calR , \calE (t) ) $, where $\calE(t)=\{(e_1, t_1), \ldots, (e_{n}, t_n) \}$ is the ordered set of historical events till time $t$ with $e_n \in \calH$ and $n$ is number events occurred and $\calR$ is the set of relation.  Here, each hyperedge $h \in \calH$  is formed of a set of recursive hyperedges, $h = h^\ell= \{ (h_{1}^{\ell-1}, r_{1}^{\ell-1}), \ldots ,(h_{k^{\ell-1} }^{\ell-1}, r_{k^{\ell-1}}^{\ell-1} ) \}$ of depth $\ell$. Here, $r_{i}^{\ell-1} \in \calR $ is the relation of hyperedge $h_{i}^{\ell-1}$ with respect to hyperedge $h^\ell$, and $h_{i}^{\ell-1} \in \calH^{\ell-1} $. Here, $k^{\ell-1}$ is the number of relational groups, or depth ${\ell-1}$ hyperedges in hyperedge $h$, and $\calE(t_a, t_b)$ is the ordered set of events observed during the interval $[t_a, t_b]$.

Using the above definition, we can represent a \textit{Directed Temporal Hypergraph} by using relations indicating the source and destination groups, here $|\calR|=2$. Furthermore, including the node type attribute, we can extend this definition to a \textit{Bipartite Temporal Hypergraph} where the nodes set is a union of two types, $\calV = \calU^r \cup \calU^\ell$. Here, $\calU^r$ and $\calU^\ell$ are the two sets of nodes, and all interactions are between them. To use this framework for \textit{ Temporal Knowledge Graph} (TKGs), where interactions are represented as a triplet of the subject entity $v^s$, object entity $v^o$, and relation $r$ between them, $h  = (v^s, r, v^o)$). Here, $v^s, v^o \in \calV$. This can be framed as a Multi-relational Recursive Hypergraph with $h = \{ (h^{0}_0, r^{0}_1), (h^{0}_1, r^{0}_2) \}$. Here, $r^0_1 =  r, r^0_2 = r^{-1}$ is the inverse relation, $h^{0}_0 = \{v^s\}$, and $h^{0}_1 = \{ v^o \}$.

\paragraph{Problem.}
Given $\calE (t) = \{ (e_1, t_1), \ldots, (e_n, t_n)\}$ historical interactions until time $t > t_n$, we aim to forecast the next interaction $(t_{n+1}, e_{n+1})$, where $t_{n+1} > t >  t_n $. Here, we want to estimate the time $t_{n+1}$ and the type of interaction $e_{n+1}$.

This work mainly focuses on higher-order interactions in communications networks and real-world events stored in source and target entities. The higher-order interaction in communications networks is represented using a recursive hyperedge graph of depth one $h =\{ (h_{i}^{0}, r_{i}^{0}) ) \}_{i=0}^{k^0}$. Here, $\{r_{i}^{0} \}_{i=0}^{3}$ relations are used to represent the sender, recipients, and carbon-copied recipients' addresses in case of email exchange and $\{r_{i}^{0} \}_{i=0}^{4}$ represents the sender, receiver, retweeter, and retweeted nodes in Twitter network. 
The higher-order interactions in real-world events are represented using depth two hyperedges $h=\{ (h_{0}^{1},r^{1}_{0} ), (h_{1}^{1}, r^{1}_{1} ) \}$ and $h^{1}_{0/1} = \{ (h_{i}^{0}, r_{i}^{0})\}_{i=0}^{3}$. Here, $h_{0}^{1}, h_{1}^{1}$ are the two entities involved in the interactions, $r^{1}_{0}$ and $r^{1}_{1}$ are the inverse relation pair indicating the type of interactions, $\{ r_{i}^{0} \}_{i=0}^3$ are relations indicating the actor, sector, country of the entity. Figure~\ref{fig:Orderd_KG} shows an example of this type of higher-order interaction.  

\section{RRHyperTPP Model}
The probability of the occurrence of hyperedge event $h$ at time $t$ is defined as follows,
$
    \rmP_h(t) = \lambda_{h} (t) \rmS_h (t^p_h, t).
$
Here, $ \lambda_{h} (t) $ is the conditional intensity function of the temporal point process, $\rmS_h (t^p_h, t)$ is the survival function that denotes the probability that the event does not occur during the interval $[t^p_h, t]$,  
$
    \rmS_h (t^p_h, t) =  \exp{ \left( -\int_{t^p_h}^t \lambda_h (\tau) \ud \tau  \right) }.
$
In this, $t^p_h < t$ is the time of the last occurrence of hyperedge $h$ before time $t$, and it is initialized to zero for all the hyperedges at the beginning. The complete likelihood for observing $\calE(T) = \{ (e_1, t_2), \ldots, (e_{N}, t_{N})\}$ in the interval $[0, T]$ can be written as, 
$
    \rmP (\calE(T)) = \prod_{n=1}^{N} \rmP_{e_n} (t_n)  \prod_{h \in \calH} \rmS_{h} (t^{\ell}_{h}, T). 
$
Here, $t^{\ell}_h$ is the last occurrence of hyperedge $h$ with $t^{\ell} = 0$ for $h \in \calH$ that is not observed in the training data. 
\begin{align}
     \rmP (\calE(T)) &=  \prod_{n=1}^{N} \lambda_{e_n} (t_n) \rmS_{e_n} (t^p_{e_n}, t_n) \prod_{ h \in \calH}  \rmS_{h} (t^{\ell}_{h}, T), \nonumber \\ 
       &=  \prod_{n=1}^{N} \lambda_{e_n} (t_n) \prod_{h \in \calH} \rmS_{h}(0, T).
\end{align}
For learning the parameters of conditional intensity $\lambda_h(t)$, we calculate the loss by taking a negative log-likelihood as shown below, 
\begin{align}\label{eq:negative_loglikelihood}
    \calN \calL \calL = - \sum_{n=1}^{N} \log \lambda_{e_n} (t_n)  + \sum_{h \in \calH} \int_{0}^T \lambda_h (t) \ud t .
\end{align}
However, doing maximum likelihood estimation (MLE) by minimization of the above loss function is computationally expensive due to the summation over all possible hyperedges in the survival function, which have an exponential number of possibilities,  and also due to the integral over the entire duration of observation $\int_0^T (. )$. So, to avoid these difficulties in computing the likelihood, we use the noise contrastive estimation technique of learning multivariate temporal point process~\citep{MeiEtAL:2020:NoiseContrastiveEstimationForMultivariatePointProcesses} as explained in Section \ref{sec:learning_and_inference}.

Furthermore, defining separate conditional intensity functions for each hyperedge will lead to difficulty in learning, as the number of parameters will increase linearly to the number of hyperedges, which is exponential on the order of the number of nodes, as explained earlier. It leads to overfitting while training and poor generalization in the test set. Hence, we define the conditional intensity function as a positive function of dynamic embeddings nodes involved in it, $\lambda_h (t)  = f (h; \bfV(t), \bfR  )$.  Here, $\bfV (t)$ is the dynamic node embedding of nodes at time $t$, and $\bfR$ is the relation embeddings. In this model, the number of parameters will be linear to the number of nodes, as hyperedges that share nodes will share the respective parameters.  Sections \ref{sec:dynamic_node_representation} and \ref{sec:hyperedge_link_prediction} describe the architecture used to implement dynamic node representation and conditional intensity function, respectively. Figure \ref{fig:appendix:rrhypertpp_mode} in Appendix \ref{sec:appendix:more_examples} contains the block diagram of the model.

\subsection{Learning and Inference}
\label{sec:learning_and_inference}

An alternate approach to MLE for learning parameters is to use noise contrastive estimation (NCE). To achieve this, we simulate  $N_q$ noise streams $\{ \mathcal{E}^{i} ( T)\}_{i=1}^{N_q}$ from the noise distribution  $\rmQ(T) $ in addition to observed data $\mathcal{E} (T) $.
The complete loss function for the noise contrastive estimation can be written as follows, 
\begin{align}\label{eq:nce_loss_complete}
    \calL_{NCE} &= -\bfE_{ \calE(T) \sim \rmP(.), \{\calE^{i} (T) \}_{i=1}^{N^q} \sim \rmQ(.) }\Bigl[  \sum_{  \mathcal{E}( t, t + \ud t) =\{(h,t) \}   } \nonumber \\ &\log{  \frac{\lambda_h (t) }{ \underline{\lambda_h} (t) }    }   
    + \sum_{i=1}^{N^q}  \sum_{\calE^{i}( t, t + \ud t) = \{(h,t) \} } \log{  \frac{\lambda_{h}^q (t)  }{  \underline{\lambda_h} (t)   } }   \Bigr].
\end{align}
Here, $\underline{\lambda_h} (t)  = \lambda_h (t)  + \lambda_{h}^q (t)  N^q $. In the above equation, we contrast the samples from the true distribution, the observed data, to the $N^q$ independently sampled noise stream. Appendix \ref{sec:appendix:proof_details} contains the details of the derivation.

Additionally, to direct the gradients toward better model parameters, we add a classification-based noise contrastive loss at the time of the event as follows, 
\begin{align} \label{eq:supervised_nce_loss}
    \calL_{NCE}^{s} = \sum_{(e_n, t_n) \in \calE(T)} \log{ \frac{\lambda_{e_n}(t)}{ \lambda_{e_n}(t) + \sum_{h \in \calH^{neg}_{e_n} } \lambda_h (t) }}. 
\end{align}
Here, $\calH^{neg}_{e_n}$ are the negative samples generated by negative sampling for the hyperedge $e_n$. Then, the combined loss can be written as follows,
\begin{align} \label{eq:combined_nce_loss}
    \mathrm{Loss} = \calL_{NCE} + \alpha \calL_{NCE}^{s} .
\end{align}
Here, $\alpha$ is a hyperparameter. 


\begin{algorithm}[H]
\caption{Training procedure of RRHyperTPP}
\label{alg:training}
\begin{algorithmic}
\STATE \textbf{Input:} $\calG(T)$.
\WHILE{not convergence} 
\STATE $\calH^{neg} \sim \mathrm{CorruptionModel}(\calH )$.
\STATE $\calH^c = \calH \cup \calH^{neg}$.
\STATE Set $t=t_0=0$, $n=1$, $\calL_{NCE} =0$.
\FOR{$n \leq N$}
\WHILE{$  t < t_n$}
\STATE Sample $\lambda^q (t) \sim \frac{1}{Nw}\sum_{i=1 }^N  \calK ( \lambda^q-\frac{1}{t_{i} - t_{i-1}}; w) $.
\STATE Sample $\Delta t  \sim \mathrm{exponenetial}(N^q \lambda^q (t) )$.
\STATE Sample $h \sim \mathrm{Uniform}(\calH^c)$.
\STATE Next event time  $t= t + \Delta t $.
\STATE $\mathrm{Loss} = \mathrm{Loss}  - \log{  \frac{\lambda_{h}^q (t)  }{  \underline{\lambda_h} (t)   } }$ 
\ENDWHILE
    \STATE $\calH^{\mathrm{neg}}_{e_n} \sim \mathrm{CorruptionModel}(e_n ) $ 
    \STATE $\mathrm{Loss}  = \mathrm{Loss}   - \log{  \frac{\lambda_{e_n} (t)  }{  \underline{\lambda_{e_n}} (t)   } }$ 
    \STATE $\mathrm{Loss}  = \mathrm{Loss}   - \alpha \log{\frac{\lambda_{e_n} (t)}{\lambda_{e_n} (t) + \sum_{h \in \calH^{\mathrm{neg}}_{e_n} } \lambda_h (t) }}  $
\STATE $n=n+1$.
\STATE If $ n \mod  B =0$, Update the model parameters by AdamW Optimizer and set $\mathrm{Loss}  = 0 $.
\ENDFOR
\ENDWHILE
\end{algorithmic}
\end{algorithm}


\subsection{Noise Distribution}
\label{sec:noise_distribution}
Here, we propose a method for simulating noise sequences by modeling the noise distribution $\rmQ(.)$.
We model the rate of events $\lambda^q$ using a stochastic process independent of the event type and time to reduce the computational complexity. The probability density function of this process is learned from the observed event times in the training data $\calE(T)$ using kernel density estimation \citep{SilvermanEtAL:1986:DensityEstimationForStatisticsAndDataAnalysis},
 $\lambda^q   \sim \frac{1}{Nw}\sum_{i=1 }^N  \calK \left(\lambda^q - \frac{1}{t_{i} - t_{i-1}}; w\right). $
 Here, $\calK$ is the kernel, and $w$ is the bandwidth of the kernel. We use the Gaussian kernel, and bandwidth is learned from the dataset. This will allow for a closed-form sampling of noise event times. Then, we generate the type of event by uniformly sampling from a set of candidate hyperedges $\calH^c$. Then $\lambda^q_h = \frac{1}{|\calH^c|}\lambda^q  $
 and 
 $\rmQ ( \calE^{i}( t, t + \ud t)=\{(h, t)\}  ) = \frac{1}{|\calH^c|}\lambda^q   \exp{ ( - \lambda^q  ( t - t_n)  )  }.$
The set of candidate hyperedges, denoted as $\calH^c$, is formed by combining the true hyperedges observed in the training data with the negative hyperedges generated through the corruption of observed hyperedges. For depth one hyperedge datasets, we expand the hyperedge as a tuple of node and relation pairs in ascending order. Then, we learn the categorical distribution of each position condition on all the previous relations, along with an end state. For each positive hyperedge, we first sample the relations till the end state is observed. Subsequently, we randomly populate each position in the negative hyperedge with nodes. This is achieved by populating each position with nodes from the true hyperedge or by randomly selecting nodes, ensuring no repetition if they share the same relation value. The process ensures that at most half of the nodes in the negative hyperedges are from the true hyperedge, with the remainder filled randomly, thus maintaining diversity in the negative samples. For depth two hyperedge datasets, we randomly corrupt any one of the depths one hyperedge within them by the above procedure to generate negative hyperedges. Here, for each observed hyperedge $h \in \calH$, we generate $N^e= |\calH^{\mathrm{neg}}_h|$ negative hyperedges. For generating $N^q$ noise streams, we multiply the noise distribution intensity function by $N^q$, $\lambda^q (t) N^q$, and simulated samples. This scaling will not affect NCE loss Equation \ref{eq:nce_loss_complete}, as all the noise streams have the same intensity values.

Algorithm \ref{alg:training} summarizes the entire training procedure of the model. The $\mathrm{CorruptionModel}(.)$ is the negative hyperedge generation function, and $\mathrm{Uniform}(\calH^c)$  uniformly samples a hyperedge from the candidate hyperedges, $\calH^c$. In the following section, we explain the dynamic node representation learning used to parameterize the conditional intensity function of the model. In our implementation, we use the same negative samples to generate the noise stream, and in the supervised noise contrastive loss in Equation \ref{eq:supervised_nce_loss}, $\calH^n = \cup_{h \in \calH} \calH^{\mathrm{neg}}_{h}$.

\section{Dynamic Node Representation}
\label{sec:dynamic_node_representation}

The dynamic node representation of the nodes $\bfV \in \bbR^{|\calV| \times d}$ in the networks are learned using a continuous time recurrent-neural network model using two stages, (i) Node Update and (ii) Drift,  for node evolution. 
\subsection{Node Update} \label{sec:node_update}

This stage is used to update the representation of a node $v$ when it is involved in an interaction $h=\left\{ (h_{1}^{\ell-1}, r_{1}^{\ell-1}), \ldots, (h_{k^{\ell-1} }^{\ell-1}, r^{\ell-1}_{k^{\ell-1}} ) \right\}$ at time $t$. The update equation uses a recurrent neural network-based architecture that updates the node's previously stored embedding based on the features of interaction calculated as follows, 
\begin{align} \label{eq:update_equation}
    \bfi^{h}_{v} &=  \MLP \left( \left[  \bfd_{v}^h;   \bfv(t^{-});  \bmPsi(t - t^p_{v}) \right]\right)   \nonumber \\ 
    \bfv (t^{+}) &= \RNN \left( \bfv(t^p_{v}) , \bfi^{h}_{v} \right).
\end{align}
Here,  $v \in h^0, (h^0,r^0) \in  \ldots \in  h$, $\MLP$ is a two-layer multi-layer perceptron. Appendix \ref{sec:appendix:mlplayers} explains the architecture of the MLP layer.
%
Here, $\bfd_{v}^h \in \bbR^d$ is the dynamic embedding calculated as a function of embeddings of other nodes involved in the interaction, $\bfd_{v}^h = f_{dyn}(h; \bfV(t), \bfR )$. The architecture of $f_{dyn}(\cdot)$ is shared with the hyperedge link predictor explained in Section \ref{sec:hyperedge_link_prediction}.  The second term $\bmPsi (t - t^p_{v}) \in \mathbb{R}^d$  is the Fourier features for the duration since an event was observed on $v$, ($t-t^p_{v}$).  Here, $t^p_{v}$ is the recent time node $v$ is involved in an interaction, and Fourier time features \cite{XuEtAL:2020:InductiveRepresentationLearningOnTemporalGraphs} are defined as  
$
        \bmPsi (t) = [\cos(\omega_1 t + \phi_1) , \ldots, \mathrm{cos}(\omega_d t + \phi_d)  ]
$, where $\{\omega_i \}_{i=1}^d$, and  $\{\phi_i \}_{i=1}^d$ are learnable parameters. When $v$ is involved in multiple positions in hyperedge $h$, an average of $\bfi^h_{v}$ is taken to update the node embedding.


\subsection{Drift} \label{sec:drift}
This stage is used to model the evolution of the nodes during the inter-event period and avoid the staleness of node embeddings due to not observing any event for a long duration \cite{KazemiEtAL:2020:RepresentationLearningForDynamicGraphASurvey}. 

\paragraph{Time Projection.}  Here, we project the node embeddings from the node update stage based on the elapsed time since the last update as follows, 
\begin{align} \label{equation:jodie_drift}
\bfv (t) = ( 1+ \bfW_t(t - t^p_v) ) \circ \bfv (t^p_v).   
\end{align}
Here, $\bfW_t \in \bbR^{d \times 1 }$ is a learnable parameter, and $\circ$ denotes the Hadamard product. This version of the embedding projection is introduced in the work JODIE \cite{KumarEtAL:2019:PredictingDynamicEmbeddingTrajectoryInTemporalInteractionNetworks}. 

\paragraph{Time Embeddings.} Here, we use Fourier time features to model the evolution of the nodes during the interevent time as follows, 
\begin{align} \label{equation:fourier_drift}
    \bfv (t) = \mathrm{tanh} (\bfW_s \bfv (t^p_v) + \bfW_t \bmPsi(  t - t^p_v)) .
\end{align}
Here, $\bfW_s, \bfW_t \in \bbR^{ d \times d}$ are learnable parameters. This form of drift stage is followed in previous work by \citet{GraciousEtAL:2023:DynamicRepresentationLearningWithTemporalPointProcessesForHigherOrderInteractionForecasting}.
\paragraph{Neural ODE.} The node embedding evolution is modeled using a neural Ordinary Differential Equation (ODE) ~\citep{ChenEtAL:2018:NeuralOdinaryDifferntialEquations} as follows,
\begin{align} \label{equation:ode_drift}
    \bfv (t)  = \bfv (t^p_v) +  \int_{t^p_v}^t f_{\nabla \bfv } \left( \bfv (t), \bmPsi(t - t^p_v) \right)  \ud t.
\end{align}
Here, $f_{\nabla \bfv }$ is the gradient of the node embedding $\bfv(t)$ implemented using a two layer multi-layer neural network.  Appendix \ref{sec:appendix:mlplayers} explains the architecture of the MLP layer.

\section{Hyperedge Link Prediction}
\label{sec:hyperedge_link_prediction}
Given the node embeddings and relation embeddings of a hyperedge, we use the query-key-value attention mechanism used in transformer models \citep{VaswaniEtAL:2017:AttentionIsALLYouNeed} for hyperedge link prediction. For this, we first expand hyperedges at depth $0$ as a tuple of node and relation pair and create an expanded depth one hyperedge $\barh^0=\{ (v, r^0 ) \}_{ v  \in h^0, (h^0,r^0) \in h^1}$.  Here, the relation is repeated for each node in the hyperedges. Then we create embeddings for query, key, and value, $\{   \bfv_i^{\uq}(t)= \bfv_i^{\uv}(t) =  [\bfv_i (t); \bfr_{i}^0] \} \forall (v_i, r_i^0) \in \barh^0 $, then we calculate the dynamic hyperedge embedding (here, dynamic is used to indicate the hyperedge dependent embeddings $\bfd^{*}_{*}$'s) for each node as follows, 
\begin{align} \label{eq:dynamic_embeddings_depth_0}
    \bfd_{v}^{\barh^0}  = \mathrm{MHAtt}( \{ \bfv^{\uq}(t) \},  \{ \bfv_j^{\uk}(t) \}_{j=1}^{\bark^{0} }, \{ \bfv_j^{\uv}(t) \}_{j=1}^{\bark^{0} } ).
\end{align}
Here, $\mathrm{MHAtt}(\cdot)$ is the MultiHeadAttention architecture proposed by \citet{VaswaniEtAL:2017:AttentionIsALLYouNeed}. Based on these, we create the depth one hyperedge embeddings by averaging out all the above node embeddings  $\bfh^1_i = \sum_{v \in \barh^0_i} \frac{\bfd^{\barh^0_i}_{v}}{|\barh^0_i|} $. Then, for the hyperedge at depths $2 \leq \ell \leq l$, we apply the following self-attention layer with query, key, and value  as $\{   \bfv_i^{\uq}(t)= \bfv_i^{\uk}(t) = [\bfh_i^{\ell-1} (t); \bfr_{i}^{\ell-1}] \}, \text{for all} (h_i^{\ell-1}, r_i^{\ell-1}) \in h^\ell $,
\begin{align} \label{eq:dynamic_embedding_higher_depth}
      \bfd_{h_i^{\ell-1} }^{h^\ell}  &= \mathrm{MHAtt}\left( \left\{ \bfv^{\uq}_i(t) \right\},  \left\{ \bfv_j^{\uk}(t) \right\}_{j=1}^{k^{\ell-1} }, \left\{ \bfv_j^{\uv}(t) \right \}_{j=1}^{k^{\ell-1} } \right), \nonumber \\
      \bfh^\ell &=  \sum_{ (h_i^{\ell-1}, r_i^{\ell-1} )\in h^\ell} \frac{\bfd_{h_i^{\ell-1} }^{h^\ell}}{| h^{\ell}|} .
\end{align}
Then conditional density is parameterized by the difference between the hyperedge embedding and dynamic hyperedge embedding of the $\ell-1$th layer as follows, 
\begin{align} \label{eq:conditional_intensity_function}
     o_i^h &= \bfW_o (\bfh_i^{\ell-1}  - \bfd_{h_i^{\ell-1} }^{h^\ell} )^2 + b_o  \nonumber \\ 
    \lambda_h(t) &= \mathrm{Softplus} \left( \sum_{ (h_i^{\ell-1}, r_i^{\ell-1} )\in h } o_i^h \right) .
\end{align}
The dynamic hyperedge embedding used in the interaction update is defined as follows: $$\bfd^h_v = [\bfd^{h^0}_v; \bfd^{h^1}_{h^0}; \ldots; \bfd^{h^\ell}_{h^{\ell-1}}; \bfh^{\ell}],$$
here $v \in h^0, h^0 \in h^1, \ldots, h^{\ell-1} \in h$.

\begin{table}
    \centering
    \small 
    \setlength{\tabcolsep}{3pt}
     \begin{tabular}{lccccccccc}
        \toprule
            \textbf{Datasets}  &    $|\calV|$  & $|\calE(T)|$ & $|\calR|$ & $T$ & depth ($l$)  \\
            \midrule 
            \textbf{Enron} & 98  & 10,355 & 3 & 10,443.0 & 1 \\
            \textbf{Twitter} &  2,714  & 52,383 &  4 & 54,028 & 1
            \\ \textbf{Boston} & 2,400  & 20,070 & 4 &  20,069 & 1 \\
            \textbf{Obama}  & 1,721  & 22,690 & 4 &  22,689 & 1 \\
            \textbf{Pope}  & 6,648 &  67,779 & 4 & 67,778 & 1 \\
            \textbf{Cannes}  & 672 &  9,078 & 4 & 9,077 & 1 \\
            \textbf{ICEWS-India} & 1,066 & 86,609  & 400 & 364 & 2 \\
            \textbf{ICEWS-Nigeria} & 894 & 69,433 & 360 & 715 & 2  \\
        \bottomrule
    \end{tabular}
    \caption{Temporal Multi-Relational Recursive Hypergraphs datasets and their vital statistics.}
    \label{tab:datasets_temporal}
\end{table}

\begin{table*}
\centering
\small 
\setlength{\tabcolsep}{1.pt}
\scalebox{0.8105}{
\begin{tabular}{lrrrrrrrrrrrrrrrr }
\toprule
\multirow{2}{*}{\textbf{Methods}}  & \multicolumn{2}{c}{\textbf{Enron}} &  \multicolumn{2}{c}{\textbf{Twitter}}  & \multicolumn{2}{c}{\textbf{Boston}}  &\multicolumn{2}{c}{\textbf{Obama}} & \multicolumn{2}{c}{\textbf{Pope}}  & \multicolumn{2}{c}{\textbf{Cannes}}   \\
\cmidrule{2-13} 
 & \textbf{AUC} & \textbf{MAE} & \textbf{AUC} & \textbf{MAE} & \textbf{AUC} & \textbf{MAE} & \textbf{AUC} & \textbf{MAE}   & \textbf{AUC} & \textbf{MAE} & \textbf{AUC} & \textbf{MAE} \\
\midrule
TGN\textsuperscript{a} & 85.8 $\pm$ 2.9 & NA & 97.0 $\pm$ 0.5 & NA & 89.3 $\pm$ 1.5 & NA & 93.0 $\pm$ 1.0 & NA & 92.5 $\pm$ 1.2 &  NA &  87.6  $\pm$ 2.4 & NA \\
HGDHE\textsuperscript{b} & 87.6  $\pm$ 0.8 & 3.6 $\pm$ 0.1 &	88.0 $\pm$  1.1 & 26.26 $\pm$ 0.5 &	75.3 $\pm$ 1.3 & 49.16 $\pm$ 5.2 &			82.0 $\pm$ 0.1 &	 29.8 $\pm$ 0.8 &		79.1 $\pm$  0.3 & 33.6 $\pm$ 1.2  &			79.1 $\pm$ 1.0  & 27.9 $\pm$ 2.0 \\
\midrule
RRHyperTPP-N &  91.0 $\pm$ 0.9 &  3.5 $\pm$ 0.0  &  94.2 $\pm$ 0.6   &  $\bm{21.9 \pm 0.4}$    &  80.8 $\pm$ 0.2  & $\bm{39.9 \pm 0.1}$  &  85.2 $\pm$ 0.4    & $\bm{27.6 \pm 0.6}$    & 82.5 $\pm$ 0.5   &  $\bm{31.9 \pm 0.3}$   &  81.9 $\pm$ 1.0 & $\bm{25.4 \pm 1.8}$       \\

\midrule
RRHyperTPP-j &  93.3 $\pm$ 0.7 & $\bm{3.3 \pm 0.2}$ & 98.5 $\pm$ 0.1 & 30.6 $\pm $1.8 & $\bm{89.2 \pm 0.1}$ &  45.7 $\pm$ 0.4 & $\bm{93.5 \pm 0.5}$ & 32.5 $\pm$ 0.4  & 94.2 $\pm$ 0.4 & $\bm{36.2 \pm 1.0}$ & 92.1 $\pm$ 0.4  & 25.6 $\pm$ 1.5 \\
RRHyperTPP-f & $\bm{93.4 \pm 0.3}$ & 3.5 $\pm$ 0.2 & $\bm{98.4 \pm 0.1}$ & 32.3 $\pm$ 0.5 & 89.2 $\pm$ 0.2 & 45.1 $\pm$ 0.4 & 92.9 $\pm$ 0.3  & 31.8 $\pm$ 0.3   & $\bm{93.9 \pm 0.3}$ & 37.6 $\pm$ 0.2 & 89.3 $\pm$ 0.4 & 25.2 $\pm$ 0.9  \\
RRHyperTPP-o & 93.5 $\pm$ 0.2 & 3.7 $\pm$ $0.1$ & 98.3 $\pm$ 0.1 & 30.1 $\pm$ 0.9 & 88.0 $\pm$ 1.7 & 46.3 $\pm$ 3.1 & 92.3 $\pm$ 1.0 & 35.6 $\pm$ 2.0 & $\bm{94.0 \pm 0.7}$ & 35.4 $\pm$ 1.2 & $\bm{91.2 \pm 0.7}$ &  28.6 $\pm$ 0.6\\
\bottomrule
\end{tabular}}
\caption{Performance on interaction type and duration prediction tasks on hyperedges of depth one. Here, interaction type prediction is evaluated using AUC in $\%$, and interaction duration prediction is evaluated using MAE. The proposed model RRHyperTPP beats baseline models in almost all settings. Here, -j, -f, and -o indicate the drift stage used. Citations: \textsuperscript{a} \citep{RossiEtAL:2020:TemporalGraphNetworksForDeepLearningOnDynamicGraphs}, \textsuperscript{b} \citep{GraciousEtAL:2023:DynamicRepresentationLearningWithTemporalPointProcessesForHigherOrderInteractionForecasting} }
\label{tab:dynamiclinkprediction_depthone}
\end{table*}
\begin{table}
\centering
\small 
\setlength{\tabcolsep}{1pt}
\scalebox{0.95}{
\begin{tabular}{lrrrr }
\toprule
\multirow{2}{*}{\textbf{Methods}}  & \multicolumn{2}{c}{\textbf{ICEWS-India}} &  \multicolumn{2}{c}{\textbf{ICEWS-Nigeria}}  \\
\cmidrule{2-5}
 & \textbf{AUC} & \textbf{MAE} & \textbf{AUC} & \textbf{MAE} \\
\midrule
RRHyperTPP-j & $\bm{58.7 \pm 0.7}$ & $\bm{0.99 \pm 0.0}$  & 61.0 $\pm$ 0.3 & $\bm{0.98 \pm 0.0}$ \\
RRHyperTPP-f & 58.9 $\pm$ 0.2 & 0.75 $\pm$ 0.2 & 60.5 $\pm$ 0.5 & 0.81 $\pm$ 0.1  \\
RRHyperTPP-o & 58.2 $\pm$ 1.4 & 0.99 $\pm$ 0.0 & $\bm{58.7 \pm 0.3 }$ & 0.97 $\pm$ 0.0\\
\bottomrule
\end{tabular}}
\caption{Performance on interaction type and duration prediction tasks on hyperedges of depth two.}
\label{tab:dynamiclinkprediction_depth_two}
\end{table}

\section{Experimental Settings and Results}

\paragraph{Datasets.}
Table \ref{tab:datasets_temporal} shows the vital statistics of the datasets used in this work. These datasets contain depth one hyperedges and are created from the works of \citet{ChodrowEtAL:2019:AnnotatedHypergraphsModelsAndApplications,ManlioEtAL:2020:UnravelingTheOriginOfSocialBurstsInCollectiveAttention,OmodeiEtAL:2015:CharacterizingInteractionInOnlineSocialNetworksDuringExceptionalEvents}. More details of the datasets and preprocessing are provided in Appendix \ref{sec:appendix:datasets}.

\paragraph{Baselines.}
TGN~\citep{RossiEtAL:2020:TemporalGraphNetworksForDeepLearningOnDynamicGraphs} a state-of-the-art pairwise interaction prediction model, HGDHE~\citep{GraciousEtAL:2023:DynamicRepresentationLearningWithTemporalPointProcessesForHigherOrderInteractionForecasting} a state-of-the-art hyperedge interaction forecasting model, and we created the model RRHyperTPP-N to show the advantage of our training strategy over negative sampling explored for training HGDHE. In RRHyperTPP-N, we use time embeddings for drift to make the model closer to the one proposed in their work. 


\paragraph{Forecasting Tasks And Evaluation Metrics.}
We use two tasks, (i) Interaction type prediction and (ii) Interaction duration Prediction, to evaluate our models' performance. In the first task, we predict which type of hyperedge occurs at time $t$ given the history. This can be estimated by finding the hyperedge with maximum conditional intensity function, $\hat{h} = \argmax_h \lambda_h(t)$. We evaluate it using the Area Under the Curve (AUC) metric \citep{Fawcett:2006:AnIntroductionToROCAnalysis} by using the $\mathrm{CorruptionModel}$ to create $N^e$ false positive samples for each true positive sample. 
For the second task, we try to predict the time at which interaction $h$ occurs from the last interaction duration of nodes in the hyperedge $t^p_h$. This is to be estimated using the condition probability density as  $\hat{t_i} = \int_{t^p_h}^{\infty} t \rmP_h (t | \calE(t^p_h))  \ud t  $.  Then, for evaluation, we use Mean Absolute Error (MAE) metric, $\text{MAE} =\frac{1}{N} \sum_{i=1}^N |\hat{t_i} - t_i | $ for all the samples in the test.

\paragraph{Parameter Settings.}
In all our experiments, we fixed the embedding dimension $d$, hidden and input dimension for the $\RNN$ at $64$, batch size $B=128$, number of noise streams $N^q=20$, number of corrupted hyperedges per true hyperedge $N^e=20$, and number heads of $\mathrm{MHAtt}$ to four. The training is done for $200$ epochs, and the model parameters that gave the least validation loss are used for testing.  For all the training, we use the AdamW optimizer \citep{SashankEtAL:2018:OnTheConvergenceOfAdamAndBeyond,IlyaEtAL:2019:DecoupledWeightDecayRegularization} of PyTorch \citep{KingmaEtAL:2015:AdamAMethodForStochasticOptimization} with learning rate set to $0.0005$. The first $50\%$ of interactions is used for training, the next $25\%$ for validation, and the rest for testing.

\subsection{Results}
\label{sec:results}
Table \ref{tab:dynamiclinkprediction_depthone} shows the performance of our models RRHyperTPP against baseline models. Here, the models that use time projection, ODEs, and time embeddings for \textbf{Drift} stage are denoted by -j, -o, and -f at the end, respectively. Here, we can also observe that no single model outperforms the rest in all the tasks.  However, theoretically, Neural ODEs should perform better than others as they have fewer assumptions when compared to Fourier and time projection-based methods. We have also done experiments on depth two hyperedges, and the results are shown in Table \ref{tab:dynamiclinkprediction_depth_two}. 

%
%
%
%
%
Moreover, when comparing our models against TGN, a state-of-the-art pairwise interaction forecasting model, we can observe an increment of  $2.9\%$, $2.3\%$, and $2.2\%$  for RRHyperTPP-j, RRHyperTPP-f, and RRHyperTPP-o, respectively, in interaction type prediction task. Hence, one can conclude that recursive hyperedge-based models perform better than pairwise link prediction models. Similarly, we compared our model RRHyperTPP-f against the HGDHE state-of-the-art hyperedge forecasting model, as both use time embeddings for the drift stage. Here, we can observe a gain of $13.6\%$ in AUC for interaction type prediction. We can observe similar improvements for other types of drift compared to the HGDHE model. 

To show the advantage of our training strategy over negative sampling-based loglikelihood approximation followed in the previous work by \citet{GraciousEtAL:2023:DynamicRepresentationLearningWithTemporalPointProcessesForHigherOrderInteractionForecasting}, we compare models RRHyperTPP-f against RRHyperTPP-N. Here, both models use the same architecture, but the former uses noise-contrastive loss for training, and the latter uses negative sampling based on approximate negative log-likelihood. There is an average improvement of $8.2\%$ in AUC for RRHyperTPP-f models compared to RRHyperTPP-N. However,  there is an average increase of $15.3\%$ in MAE for RRHyperTPP-f compared to RRHyperTPP-N. Hence, more exploration is needed to find a better noise generation strategy to achieve better performance simultaneously in both tasks. Further, RRHyperTPP-N has an improvement of $4.9\%$ in AUC and $9.8\%$ reduction in MAE over previous work HGDHE, which uses hyperedge to represent higher-order interaction. Hence, we can conclude that multi-relational recursive provides better representation for higher-order interactions than hyperedges.
\begin{figure}
    \centering
    \includegraphics[width=0.6\linewidth]{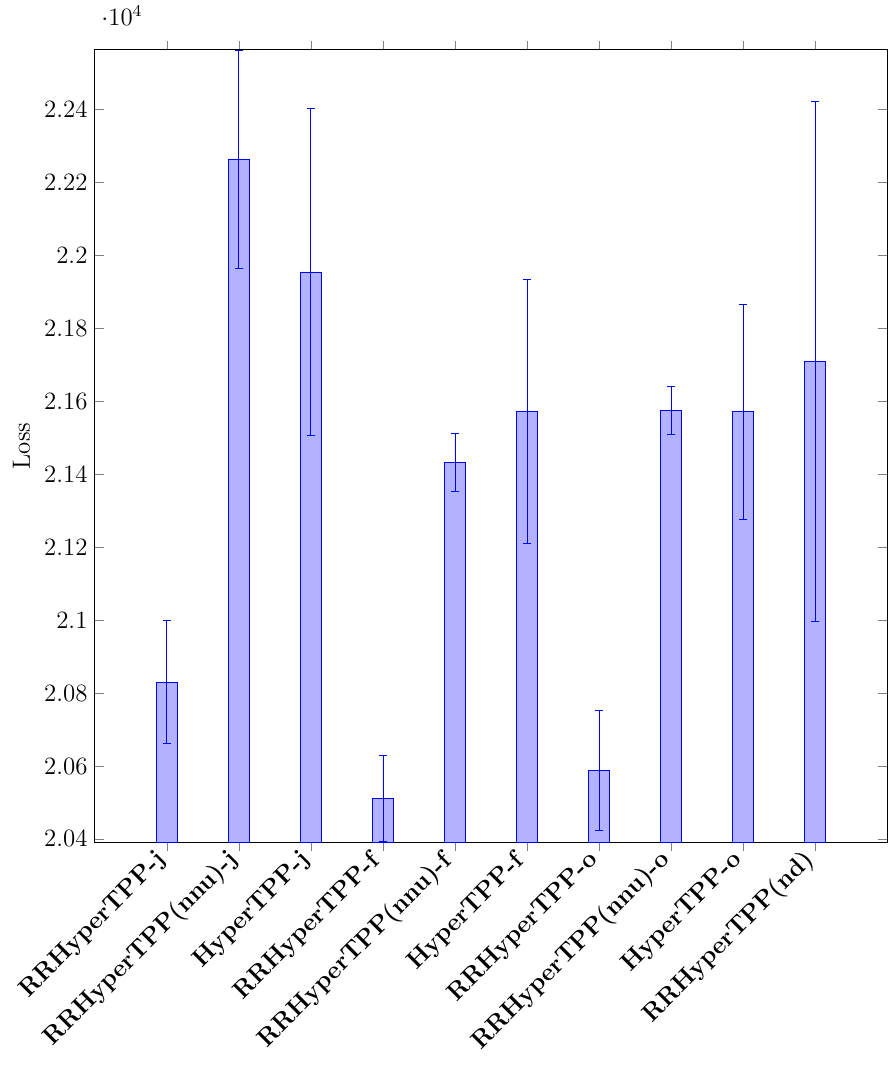}
    \caption{$\mathrm{Loss}$ for RRHyperTPP and its variants on Enron dataset.  We can observe that the proposed RRHyperTPP models perform better than their variants.}
    \label{fig:ablationstudiesenron}
\end{figure}

\paragraph{Ablation Studies.}
To evaluate the efficiency of the dynamic node representation architecture described in Section \ref{sec:dynamic_node_representation}, we conduct a series of ablation studies by systematically removing components of the model. We define the RRHyperTPP(nnu) baseline by excluding the \textbf{Node Update} stage while retaining the \textbf{Drift} stage. This results in the following variants: RRHyperTPP(nnu)-j/f/o.
%
Additionally, we create the RRHyperTPP(nd) variant by removing \textbf{Drift} stage while keeping \textbf{Node Update}. Figure \ref{fig:ablationstudiesenron} presents the performance of these models on the Enron dataset. The results indicate that the RRHyperTPP models achieve an average $5.1\%$ reduction in loss compared to the RRHyperTPP(nnu) models that exclude the \textbf{Node Update} stage. This demonstrates that incorporating the \textbf{Node Update} stage enhances performance by enabling the dynamic node representation to evolve with each new interaction. Furthermore, the RRHyperTPP(nd) model, which excludes \textbf{Drift} stage, exhibits a $4.9\%$ higher loss compared to the proposed model. This underscores the importance of the \textbf{Drift} stage in dynamic node representation, as it aids in modeling the evolution of node representations during interevent periods, as discussed in Section \ref{sec:drift}. Furthermore, to show the advantage of the recursive hyperedge link prediction in Section \ref{sec:hyperedge_link_prediction}, we created baseline models HyperTPP-j/f/o to forecast hyperedges created from the nodes involved in the interactions, $h=\{ v_i; v_i \in h^0, (h^0,r^0) \in \ldots \in h\}$. Here, we can observe a $4.8\%$ reduction in Loss for RRHyperTPP models when compared to HyperTPP-based models in interaction-type prediction tasks.  Similar trends are observed across other datasets, as detailed in Appendix \ref{sec:appendix:ablation_studies}.

\
\begin{figure}
    \centering
    \includegraphics[width=0.95\linewidth]{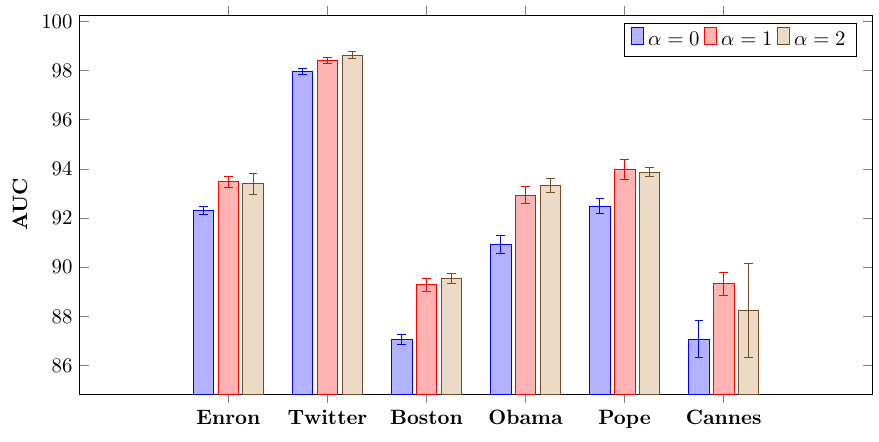}
\caption{The performance gain obtained due to the addition of classification-based noise contrastive loss mentioned Equation \ref{eq:combined_nce_loss}. Here, we can observe that models trained with supervised noise contrastive loss $\alpha>0$ have more AUC scores for interaction type prediction than models that do not ($\alpha=0$). }
    \label{fig:ablation_supervised_loss}
\end{figure}

\paragraph{Performance gain due to supervised NCE. } Figure \ref{fig:ablation_supervised_loss} illustrates the performance improvement in AUC of RRHyperTPP-f with different weights of the supervised noise contrastive loss in Equation \ref{eq:combined_nce_loss} for the interaction type prediction task. Here, it can be observed that when $\alpha=1$, there is a $1.7\%$ increase in AUC compared to $\alpha=0$. Therefore, we infer that the supervised noise contrastive loss enhances performance. Additionally, it can be noted that when $\alpha$ is increased to two, there are no significant improvements, and a performance decrement is observed for the Cannes dataset. Consequently, we can conclude that increasing $\alpha$ will not yield better performance. Furthermore, we observed no significant improvement for the interaction duration prediction task by changing $\alpha$.

\section{Conclusion}
In this work, we have established the significance of using the multi-relational recursive hyperedges for interaction modeling as it captures real-world events closely. To achieve this, we proposed a dynamic node representation learning technique and link predictor framework that captures complex relationships to forecast events. In addition, we also proposed a noise contrastive learning framework that avoids the integration calculation in the survival function and can train when the number of types of events is of exponential order. To evaluate the model, we curated six datasets of depth one and two datasets of depth two hyperedges. We observed that our model RRHyperTPP performs better than the previous state-of-the-art graph pairwise link prediction model TGN and hyperedge prediction model HGDHE. As a future direction, we plan to explore new domains of application of these models and new architecture that uses higher-order information to learn efficient node representations. In addition to this, we also plan on improving the noise generation strategy by using historical data, so the noise-generated samples will look more closely to actual data, thereby reducing the training iterations.

\section*{Acknowledgements}
The authors would like to thank the SERB, Department of Science and
Technology, Government of India, for the generous funding towards this
work through the IMPRINT Project: IMP/2019/000383.

\bibliography{aaai25,references_thesis}
\appendix
\appendix

\twocolumn[
\begin{center}
\textbf{\Large Deep Representation Learning for Forecasting Recursive and Multi-Relational
Events in Temporal Networks: Appendix }
\end{center}
 \hfill \break
 \hfill \break
 ]

\section{Notations} \label{sec:appendix:notations}

\begin{table*}
	\centering
	\begin{tabular}{|c|l|}
		\toprule
		\textbf{Notation}  &  \textbf{Definition}  \\
		\midrule 
            $\calG = (\calV, \calH)$ & HypergGraph  \\ 
            $\calV$ & Set of nodes \\
            $\calH$ & set of edges \\ 
            $\calG (t)= ( \calV, \calH, \calR , \calE (t) ) $ & Temporal multi-relational \\ 
             &  recursive hypergraph \\ 
            $\calR$ & Set of relations \\ 
            $N$ & Total number of events \\
		$t$ & Time \\
            $h$ & Hyperedge \\
		$k$ & Hyperedge size \\ 
            $\calE (t)$ &  Historical interactions till time $t$ \\
            $e_n$ & $n$th event  \\
            $n$ &  Event index \\ 
 		$t^p_v$ & Previous event time for node $v_i$ \\
            $t^p_h$ & Previous event time for hyperedge $h$ \\ 
		$\ud t$ & Derivative of time $t$ \\
		$d$ & Embedding size \\
		$v_i$ & $i$th node \\
  		$\lambda_h(t)$ & Conditional intensity \\
                & function of interaction $h$\\
            $\rmP_h(t)$ & Probability of hyperedge event $h$ 
            \\  & happening  at time $t$ \\ 
            $\rmP^{*} (\cdot) $ & $\rmP( \cdot | \calE(t))$ conditional event distribution \\ 
                & of the observed sequence \\ 
            $ \rmS_h (t^p_h, t)$ & Survival function of hyperedge \\
             & during the interval $[t^p_h, t]$ \\
             $T$ & Observation interval \\
            $\calN \calL \calL$ & Negative log-likelihood \\
            $\calL_{NCE}$ & Noise contrastive loss \\ 
            $\calL_{NCE}^s$ & Supervised noise contrastive loss \\
		\bottomrule
	\end{tabular}
        \caption{Notations-I}
	\label{tab: Notations}
\end{table*}

\begin{table*}
	\centering
	\begin{tabular}{|c|c|}
		\toprule
		\textbf{Notation}  &  \textbf{Definition}  \\
		\midrule 
            $\calC (. )$ & Power set \\  
            $\calH^\ell$ & Hyperedges at depth $\ell$ \\
            $\ell$ & Depth \\ 
            $k^\ell$ & Number of depth $\ell$ hyperedges in $h$ \\
            $h^{0}$ & Hyperedge at depth $0$ \\
            $h =\{ (h_{1}^{\ell-1}, r_{1}^{\ell-1}), \ldots ,(h_{k^{\ell-1} }^{\ell-1}, r_{k^{\ell-1}}^{\ell-1} ) \}$ & Recursive hyperedge of depth $\ell$ \\ 
            $r_i^{\ell-1}$ & Relation of hyperedge $h_{i}^{\ell-1}$ with respect to hyperedge $h^\ell$ \\
            $r^{-1}$ & Inverse relation \\ 
            $\calU^r, \calU^\ell$  &  Right nodes and left nodes \\ 
            $v^s$ & Subject entity \\ 
            $v^o$ & Object entity \\ 
 		$\ud t$ & Derivative of time $t$ \\ 
		$v_i$ & $i$th node \\
            $\rmQ(.)$ & Noise distribution \\ 
            $\calE^{i}$ & Noise stream $i$ \\ 
            $N^q$ & Number of noise streams \\
            $N^e$ & Number of negative samples per observed hyperedge \\ 
            $\varnothing$ & Null event \\ 
            $\underline{\lambda_h} (t)$ & conditional intensity function of combined event stream \\
            $\rmQ^{*} (\cdot) $ & $\rmQ( \cdot | \calE(t))$ conditional event distribution of the noise sequence \\ 
            $t^{\ell}_h $ & Last event time for hyperedge $h$ \\
            $\Delta t$ & Interevent time\\
            $\calH^c$ & Candidate hyperedges \\ 
            $\calH^{neg}_{e_n}$ & Negative hyperedges created for hyperedge $e_n$ \\ 
            $\calK$ & Kernel for KDE \\ 
            $w$ & Bandwith\\  
            $B$ & Batch size \\ 
            $\bfd_{v}^h$ & Dynamic hyperedge embedding \\ 
            $f_{dyn}$ & Function for dynamic hyperedge embedding \\
            $\bmPsi (.) $ & Fourier features \\ 
            $\{\omega_i \}_{i=1}^d$, and  $\{\phi_i \}_{i=1}^d$  & Parameters of Fourier features \\ 
            $\bfv$ & Node embeddings \\ 
            $\bfi^h_v$ & Interaction features \\
            $\bbR$ & Real space \\ 
		$\bfW_t$ & Temporal drift parameter \\
            $f_{\nabla \bfv }$ & Gradient of node embedding \\ 
            $\bfv_i^{\uq}(t), \bfv_i^{\uk}(t), \bfv_i^{\uv}(t) $ & Query, Key, Value embeddings\\
            $\barh^0$ & Expanded hyperedge at depth $0$ \\  
            $\bark^0$ & Size of expanded hyperedge at depth $0$\\
            $\bfd^{h^{\ell}}_{h^{\ell-1}_i}$ & Dynamic node embedding at depth $\ell$ for for hyperedge $h^{\ell-1}_i$\\ 
            $\bfh^\ell_i$ & Hyperedge embedding \\
            $o_i^{h}  $ & Output score for hyperedge link predictor \\  
            $\mathrm{MHAtt}(\cdot)$  & Multi-head attention layer\\ 
		\bottomrule
	\end{tabular}
        \caption{Notations-II}
	\label{tab: Notations_2}
\end{table*}

Tables \ref{tab: Notations} and ~\ref{tab: Notations_2} show the notations and their respective definitions used in this paper. We used bold lower-case letters to show vectors and bold upper-case letters to show matrices.
 \section{More Examples} \label{sec:appendix:more_examples}

 \begin{figure}
    \centering
    \includegraphics[width=0.70\linewidth]{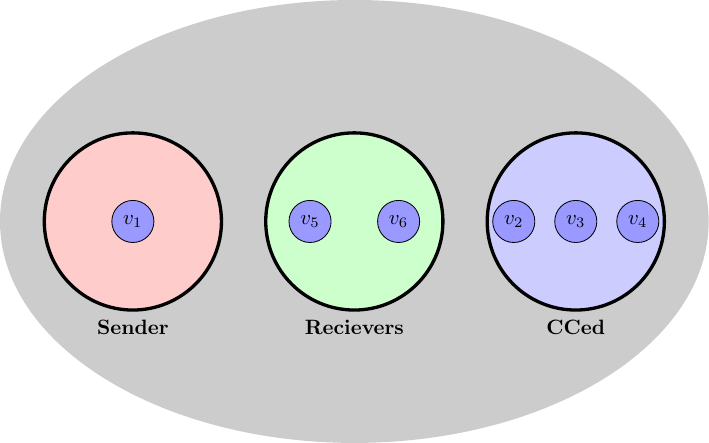}
    \caption{Email exchange between a group of people is represented as a Relational Recursive Hyperedges depth. Here, the sender involves a single person, and multiple persons are in CCed and Receiver groups.}
    \label{fig:Appendix_Orderd_email_exchange}
\end{figure}

Figure \ref{fig:Appendix_Orderd_email_exchange} shows an email exchange as a higher-order interaction of depth one. The higher-order interaction in communications networks is represented using a recursive hyperedge graph of depth one $h =\{ (h_{i}^{0}, r_{i}^{0}) ) \}_{i=0}^{k^0}$. Here, relations are used to represent $\{r_{i}^{0} \}_{i=0}^{3}$  the sender, recipients,  and carbon-copied recipients' addresses.

\section{MLP Layers} \label{sec:appendix:mlplayers}

\paragraph{$\mathrm{MLP}(\cdot)$.} The multi-layer perceptron used for interaction feature generating $\bfi^h_v \in \bbR^d$ has the following form, 
\begin{align}
    \bfi^h_v = \bfW^i_1 \mathrm{tanh} \left( \bfW^i_0 \left[  \bfd_{v}^h;   \bfv(t^{-});  \bmPsi(t - t^p_{v}) \right] \right).
\end{align}
Here, $\bfW^i_0 \in \bbR^{d*(l+3) \times d*(l+3)/2 }$, $\bfW^i_1 \in \bbR^{d*(l+3)/2 \times d } $.

\paragraph{$f_{\nabla \bfv } \left( \bfv (t), t - t^p_v  \right) $.} This is a multi-layer perceptron function used for gradient calculation for Neural ODE with the following form, 
\begin{align}
    f_{\nabla \bfv } \left( \bfv (t), \bmPsi(t - t^p_v)  \right)  = \bfW^t_1 \mathrm{tanh} \left( \bfW^t_0 \left[  \bfv(t);  \bmPsi (t - t^p_v) \right] \right).
\end{align}
Here, $\bfW^t_0 \in \bbR^{ 2d \times d}, \bfW^t_1 \in \bbR^{ d \times d}$. 

\paragraph{$\mathrm{MHAtt(\cdot)}$.} This is the MultiHeadAttention architecture proposed by \citet{VaswaniEtAL:2017:AttentionIsALLYouNeed}. In all our experiments, we fixed the number of heads to four.

\section{Datasets} \label{sec:appendix:datasets}

\begin{table*}
    \centering
    \small 
    \setlength{\tabcolsep}{3pt}
     \begin{tabular}{lccccccccc}
        \toprule
            \textbf{Datasets}  &    $|\calV|$  & $|\calE(T)|$ & $|\calR|$ & $T$ & $l$ & $\Delta t (mean)$ &  $\Delta t (max)$ &  $\Delta t (min)$ \\
            \midrule 
            \textbf{Enron} & 98  & 10,355 & 3 & 10,443.0 & 1 & 1.02 & 893.45 & 0.006 \\
            \textbf{Twitter} &  2,714  & 52,383 &  4 & 54,028 & 1 &  1.54 & 25.64 & 0.625 \\ \textbf{Boston} & 2,400  & 20,070 & 4 &  20,069 & 1 &  1.03 & 37.13 & 0.044 \\
            \textbf{Obama}  & 1,721  & 22,690 & 4 &  22,689 & 1 &  1.02 &  1628.61 & 0.021\\
            \textbf{Pope}  & 6,648 &  67,779 & 4 & 67,778 & 1 &  1.07 & 68.97 & 0.047 \\
            \textbf{Cannes}  & 672 &  9,078 & 4 & 9,077 & 1 &  1.00 & 58.02  & 0.007 \\
            \textbf{ICEWS-India} & 1,066 & 86,609  & 400 & 364 & 1 &  1.00 & 1.0  & 1.0 \\
            \textbf{ICEWS-Nigeria} & 894 & 69,433 & 360 & 715 & 1 &  1.00 & 2.94  & 0.98 \\
        \bottomrule
    \end{tabular}
    \caption{Temporal Multi-Relational Recursive Hypergraphs datasets and their vital statistics.}
    \label{tab:appendix:datasets_temporal}
\end{table*}

Table \ref{tab:appendix:datasets_temporal} shows the vital statistics of the datasets used in this work. Enron is an email exchange dataset created from the corpus \footnote{https://www.cs.cmu.edu/~enron/}. This contains a depth one hyperedge containing three relations \texttt{Sender}, \texttt{Recievers}, and \texttt{CCed} as shown in Figure \ref{fig:Appendix_Orderd_email_exchange}. The datasets Twitter, Boston, Obama, Pope, and Cannes are exchanges on the Twitter platform between different users during a period of interest. These datasets created from the following works \citet{ChodrowEtAL:2019:AnnotatedHypergraphsModelsAndApplications,ManlioEtAL:2020:UnravelingTheOriginOfSocialBurstsInCollectiveAttention,OmodeiEtAL:2015:CharacterizingInteractionInOnlineSocialNetworksDuringExceptionalEvents}. Similarly to Enron, these contain depth-one hyperedges with four relations \texttt{Sender}, \texttt{Retweet}, \texttt{Mentions} and \texttt{Replies}. ICEWS-India and ICEWS-Nigeria contain depth two hyperedges created from events stored in Integrated Crisis Early Warning System data for countries India and Nigeria~\citep{icewsdataset}. The hyperedges in these datasets are similar to one shown in \ref{fig:Orderd_KG}, except that each has a source and a target hyperedge contains two relations \texttt{Actor} and \texttt{Sector}, as the relationship country is the same for all hyperedges within the dataset. For the ICEWS-India dataset, there are 398 relations between the source and target hyperedges, while for ICEWS-Nigeria, there are 358 relations. For preprocessing, we have filtered out the nodes based on frequency, and all the timestamps start from zero and are scaled down so that the mean interevent time is around one. Here, $\Delta t (mean)$, $\Delta t (max)$, and $\Delta t (min)$ are the mean, maximum, and minimum interevent time, and $\ell$ is hyperedge depth.

\section{Proof Details} \label{sec:appendix:proof_details}

To achieve this, we discretize the training period $[0, T]$ into intervals of size $\Delta t$ and simulate $N_q$ noise streams $\{ \mathcal{E}^{i} ( t, t + \Delta t) \}_{i=1}^{N_q}$ from the noise distribution  $\rmQ(. | \calE (t) ) $ in addition to observed data $\mathcal{E}^0 ( t, t + \Delta t)  = \calE ( t, t + \Delta t) $. In the following section, we will also use the zeroth index to denote the observed events for ease of depiction. Then, the model parameters are trained to discriminate true and noise events from the candidate set $\{ \calE^{i} ( t, t + \Delta t) \}_{i=0}^{N^q}$. This can be written as follows, 
\begin{align*}
      \log&{\left( \rmP^{*} \left( \calE^0( t, t + \Delta t)  \right) \prod_{i=1}^{N^q} \rmQ^{*} ( \calE^{i}( t, t + \Delta t)   ) \right) } - \nonumber \\ 
     &\log{ \left(\sum_{i=0}^{N^q} \rmP^{*} ( \calE^{i}( t, t + \Delta t) )   \prod_{ j \neq i,j=1 }^{N^q} \rmQ^{*} ( \calE^{j}( t, t + \Delta t)  )    \right)}.
\end{align*}
Here, $ \rmP^{*} (\cdot) = \rmP( \cdot | \calE(t))$ and $ \rmQ^{*} (\cdot) = \rmQ( \cdot | \calE(t))$. However, we still need to calculate the integral $\sum_{h \in \mathcal{H}}\int_{t}^{t + \Delta t} $ in the probability density function, which is still computationally expensive. The solution is to shrink the interval to an infinitesimal width of $\ud t $ when $\lim \ud t \to 0$. Then, one can rewrite the above equation as, 
\begin{align*}
     \log&{ \left(    \rmP^{*} \left( \calE^0( t, t + \ud t)  \right) \prod_{i=1}^{N^q} \rmQ^{*} ( \calE^{i}( t, t + \ud t)   ) \right) } - \nonumber \\ &\log{ \left(\sum_{i=0}^{N^q} \rmP^{*} ( \calE^{i}( t, t + \ud t) )   \prod_{ j \neq i,j=1 }^{N^q} \rmQ^{*} ( \calE^{j}( t, t + \ud  t)  )    \right)}.
\end{align*}
Here, in the interval $[t, t+ \ud t]$, an event will be observed $\calE^0 ( t, t + \ud t) =\{(h,t) \}$  or a null event will be observed $\calE^0 ( t, t + \ud t) = \{ (\varnothing, t) \}$ with respect to the distribution $ \rmP^* \left( \calE^0( t, t + \ud t)  \right)$. So, the probability distribution should satisfy the following inequality at the interval $[t, t + \ud t ]$, 
\begin{align}
   & \sum_{h \in \calH} \rmP^*  \left( \calE^0( t, t + \ud t) = {(h_i, t)}  \right)   \nonumber \\   & +  \rmP^* \left( \calE^0( t, t + \ud t) = {( \varnothing, t )}  \right) =  1 , \nonumber \\ 
   & \sum_{h \in \calH} \lambda_{h}(t) \ud t   \nonumber \\ &   +\rmP^*  \left( \calE^0( t, t + \ud t) = { (\varnothing,t )} \right) =  1.
\end{align}
Here, $\rmP^* \left( \calE^0( t, t + \ud t) = \{(h, t)\}  \right) = \lambda_{h} \ud t $ from the temporal point process conditional intensity function definition, and 
\begin{align*}
    \rmP^* \left( \calE^0( t, t + \ud t) = { (\varnothing, 1) }  \right) = 1, \text{as $\ud t \to 0$.}   
\end{align*}
Similarly, we can define $ \rmQ^* \left( \calE^{i}( t, t + \ud t)   \right)$ using its definition of conditional intensity function. As a result of this, our objective function when there are no events in the observed stream $\calE^0 ( t, t + \ud t) = {(\varnothing, t)}$ and in the $N^q$ noise streams $\{\calE^{i }( t, t + \ud t) = \{(\varnothing, t)\} \}_{i=1}^{N^q}$ becomes $\log{ \frac{1}{N^q+1}}$. However, when event $h$ occurs in the true data stream or the noise data stream. The loss could be written as follows, 
\begin{align}\label{eq:single_point_nce}
    & \log{  \frac{\lambda_h (t) }{ \lambda_h (t)  + \lambda_{h}^q (t)  {N^q}  } } \text{ if $\calE( t, t + \ud t) = \{(h,t) \}$ else, } \nonumber \\ 
    & \text{for $i=1$ to $N^q$, } \nonumber \\ 
    & \log{  \frac{\lambda_{h}^q (t)  }{ \lambda_h (t)  + \lambda_{h}^q (t)  {N^q} } } \text{ if $ \hspace{0.1cm} \exists \hspace{0.1cm} \calE^{i}( t, t + \ud t) = \{(h,t) \} $  }. 
\end{align}

The complete loss function for the noise contrastive estimation can be written as follows, 
\begin{align}
    \calL_{NCE} &= -\bfE_{ \calE(T) \sim \rmP(.), \{\calE^{i} (T) \}_{i=1}^{N^q} \sim \rmQ(.) }\Bigl[  \sum_{  \mathcal{E}( t, t + \ud t) =\{(h,t) \}   } \nonumber \\ &\log{  \frac{\lambda_h (t) }{ \underline{\lambda_h} (t) }    }   
    + \sum_{i=1}^{N^q}  \sum_{\calE^{i}( t, t + \ud t) = \{(h,t) \} } \log{  \frac{\lambda_{h}^q (t)  }{  \underline{\lambda_h} (t)   } }   \Bigr].
\end{align}
Here, $\underline{\lambda_h} (t)  = \lambda_h (t)  + \lambda_{h}^q (t)  N^q $. In the above equation, we contrast the samples from the true distribution, the observed data, to the $N^q$ independently sampled noise stream conditioned on the observed true historical data. A more detailed explanation of this contrastive loss function and the optimality of this technique can be found in the work of \citet{MeiEtAL:2020:NoiseContrastiveEstimationForMultivariatePointProcesses}. 

\section{Model Architecture} \label{sec:appendix:model_architecture}
\begin{figure*}
    \centering
    \includegraphics[width=\textwidth]{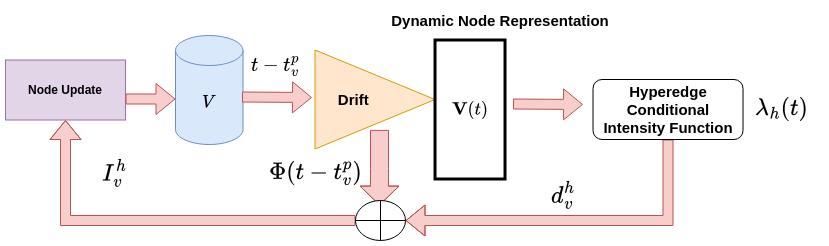}
    \caption{Deep Neural Network Architecture of RRHyperTPP: We calculate the dynamic node representation $\bfV (t)$ by using Node Update (Section \ref{sec:node_update}) and Drift (Section \ref{sec:drift}) stages. The node Update stage is used to update the node embeddings after an interaction involving that node occurs, and the Drift stage is used to model the evolution of node embedding during the interevent period. The dynamic node representation is used by the hyperedge link prediction-based decoder to infer the conditional intensity function $\lambda_h (t)$.  }
    \label{fig:appendix:rrhypertpp_mode}
\end{figure*}

Figure \ref{fig:appendix:rrhypertpp_mode} shows the block diagram of the RRHyperTPP model proposed in our work.
\section{Ablation Studies}
\label{sec:appendix:ablation_studies}
\begin{figure*}
    \centering
    \begin{subfigure}{0.3\textwidth}
        \centering
        \includegraphics[width=0.95\linewidth]{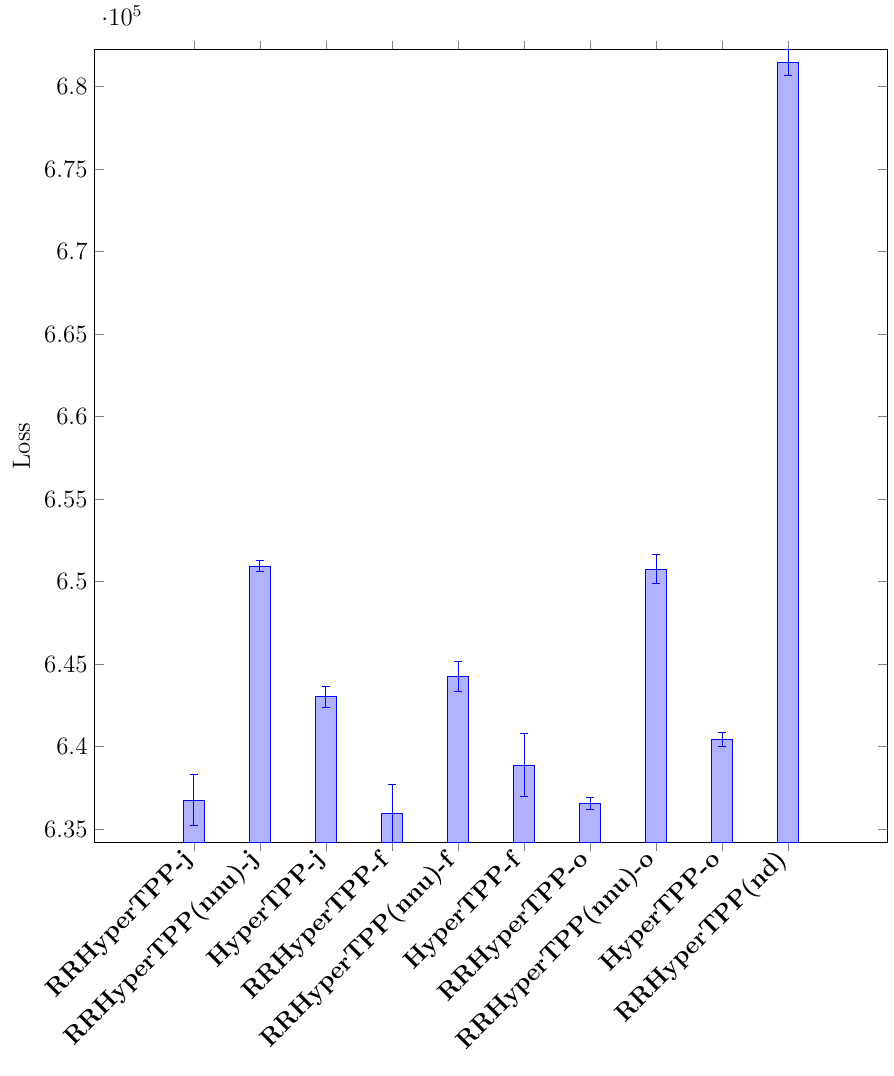}
        \caption{$\mathrm{Loss}$  for RRHyperTPP and its variants on Twitter. We can observe that the proposed RRHyperTPP models perform better than their variants.}
        \label{fig:ablationstudiestwitter}
    \end{subfigure}
    \hspace{0.001\textwidth}
    \begin{subfigure}{0.3\textwidth}
        \centering
        \includegraphics[width=0.95\linewidth]{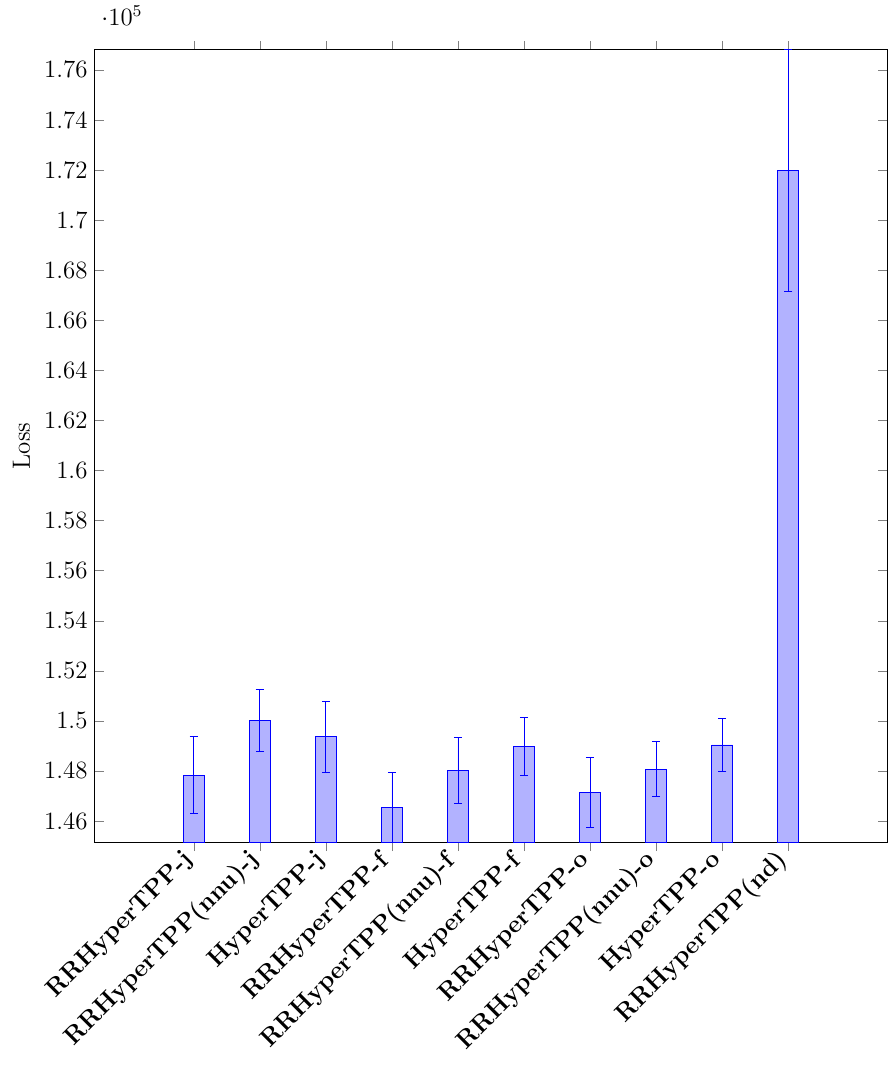}
        \caption{$\mathrm{Loss}$  for RRHyperTPP and its variants on Boston. We can observe that the proposed RRHyperTPP models perform better than their variants.}
        \label{fig:ablationstudiesboston}
    \end{subfigure}
    \hspace{0.001\textwidth}
    \begin{subfigure}{0.3\textwidth}
        \centering
        \includegraphics[width=0.95\linewidth]{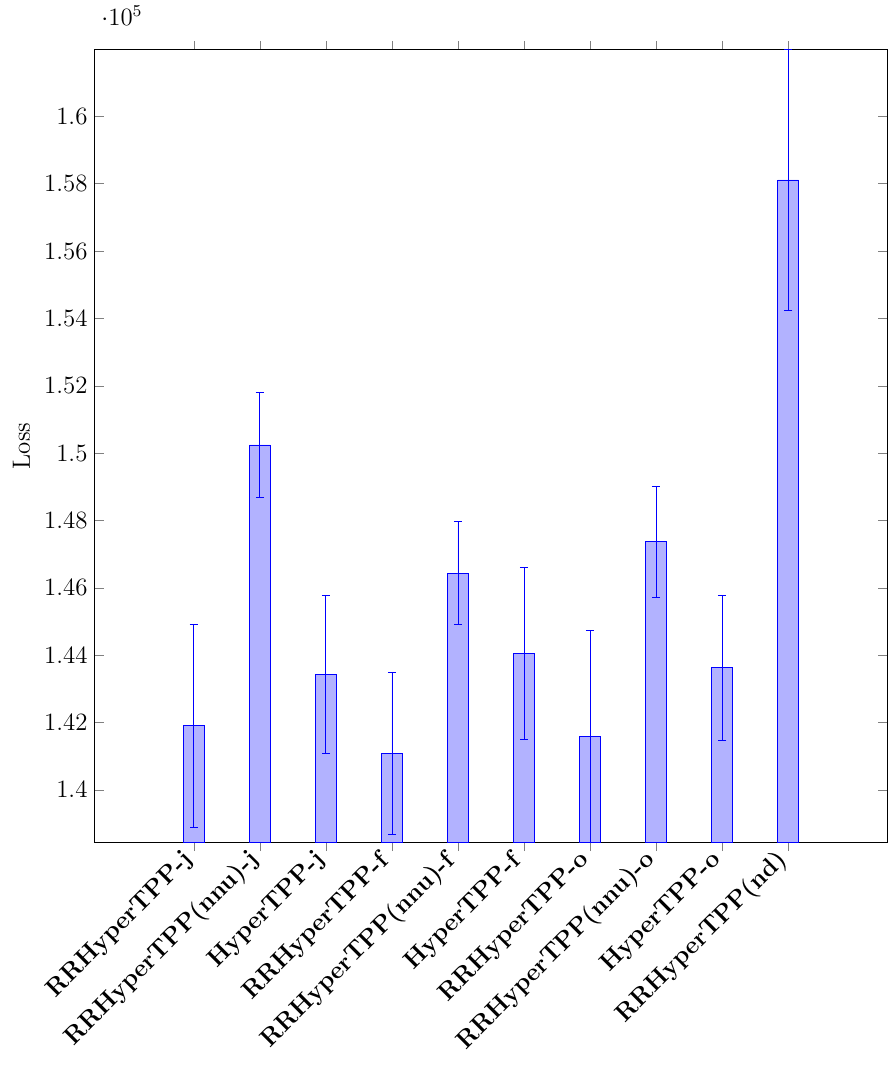}
        \caption{$\mathrm{Loss}$ for Obama dataset compared with its variants. Here, we can see that the proposed models \textsf{RRHyperTPP} models give better performance than their variants.}
        \label{fig:ablationstudiesobama}
    \end{subfigure}
    \hspace{0.001\textwidth}
    \begin{subfigure}{0.3\textwidth}
        \centering
        \includegraphics[width=0.95\linewidth]{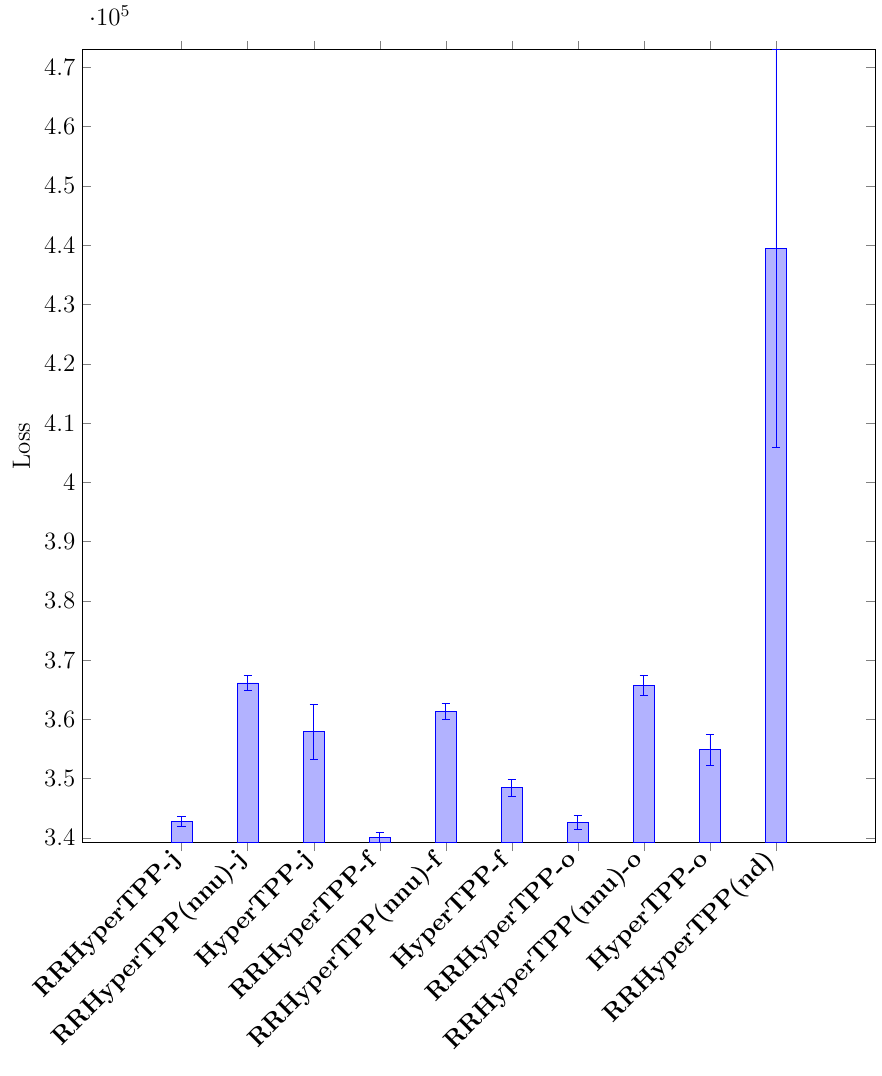}
        \caption{$\mathrm{Loss}$ for Pope dataset compared with its variants. Here, we can see that the proposed models \textsf{RRHyperTPP} models give better performance than their variants.}
        \label{fig:ablationstudiespope}
    \end{subfigure}
    \hspace{0.001\textwidth}
    \begin{subfigure}{0.3\textwidth}
        \centering
        \includegraphics[width=0.95\linewidth]{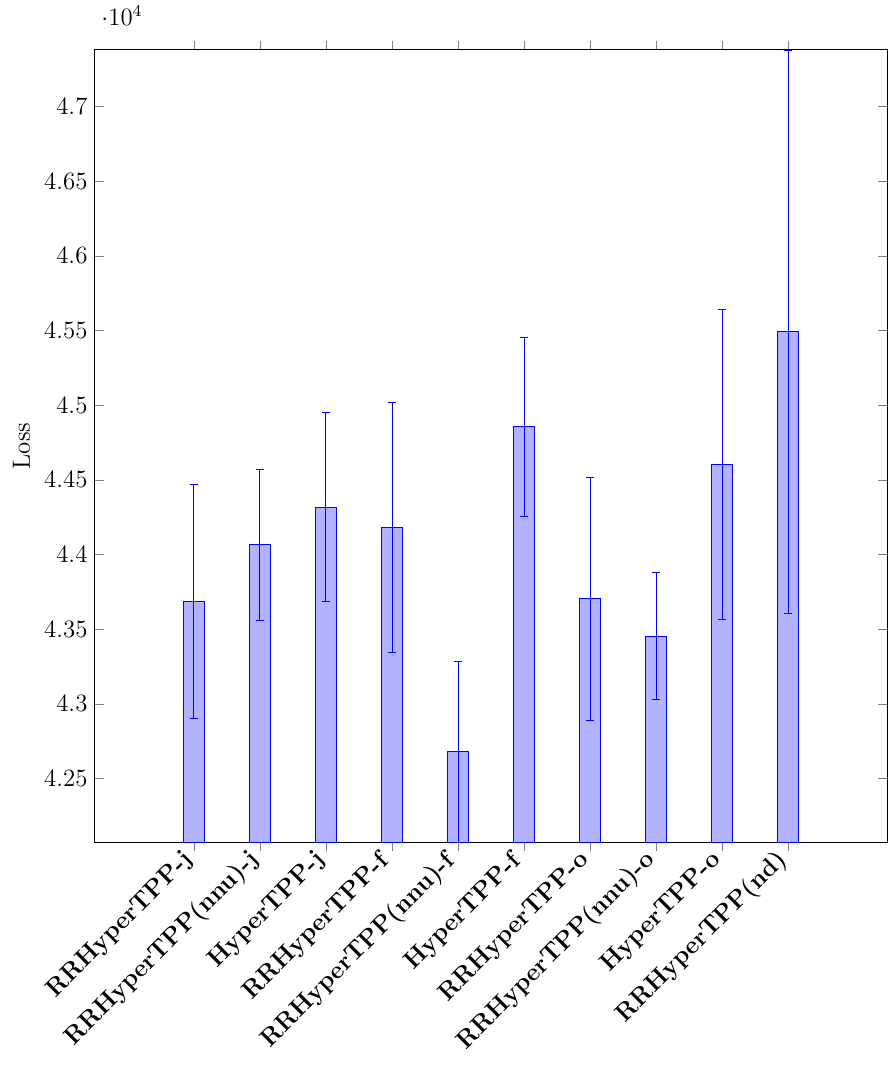}
        \caption{$\mathrm{Loss}$ for Cannes dataset compared with its variants. Here, we can see that the proposed models \textsf{RRHyperTPP} models give better performance than their variants.}
        \label{fig:ablationstudiescannes}
    \end{subfigure}
    \caption{Ablation study conducted to show efficiency of dynamic node representation architecture and multi-relational recursive hyperedge formulation for datasets Twitter, Boston, Obama, Pope, and Cannes are shown in Figures \ref{fig:ablationstudiestwitter}, \ref{fig:ablationstudiesboston}, \ref{fig:ablationstudiesobama}, \ref{fig:ablationstudiespope}, and \ref{fig:ablationstudiescannes}, respectively.}
\end{figure*}

Figures \ref{fig:ablationstudiestwitter}, \ref{fig:ablationstudiesboston}, \ref{fig:ablationstudiesobama}, \ref{fig:ablationstudiespope}, and \ref{fig:ablationstudiescannes} shows performance proposed models and its variants created to show the efficiency of dynamic node representation. Here, we can observe RRHyperTPP models reduce the loss by $1.88\%$, $1.02\%$, $4.37\%$, and $6.19\%$   compared to RRHyperTPP(nnu) models for datasets Twitter, Boston, Obama, and Pope, respectively. Hence, we can conclude that \textbf{Node Update} stage helps in learning better models as the loss calculated in the test is lesser than the models without it. Further, we can also observe that RRHyperTPP(nd) models have $6.41\%$, $1.44\%$, $10.49\%$, $22.22\%$, and 	$3.59\%$ higher loss than RRHyperTPP models for datasets Twitter, Boston, Obama, Pope, and Cannes, respectively. Hence, we can conclude that the \textbf{Drift} stage is important for building models that generalize well in the test data. Additionally, RRHyperTPP models reduce loss by $0.67\%$, $1.31\%$,	 $1.51\%$, $3.35\%$, and $1.64\%$ for datasets Twitter, Boston, Obama, Pope, and Cannes, when compared to HyperTPP models. This shows that multi-relational recursive hyperedges formulation is a more natural framework for describing these datasets.

\end{document}